\pdfoutput=1

\documentclass[11pt]{article}

\usepackage[]{acl}

\usepackage{times}
\usepackage{latexsym}

\usepackage[T1]{fontenc}

\usepackage[utf8]{inputenc}

\usepackage{microtype}

\newcommand{\task}{idiomaticity detection}

\newcommand{\glossbert}{{defBERT}}
\newcommand{\attested}{{UNATT}}
\newcommand{\translation}{{MT}}
\newcommand{\baseline}{mBERT}
\newcommand{\owt}{MT(one)}
\newcommand{\awt}{MT(all)}

\newcommand{\bnDefbert}{\glossbert{}-BN-src}
\newcommand{\bnEnDefbert}{\glossbert{}-BN-en}
\newcommand{\omwEnDefbert}{\glossbert{}-OMW-en}

\newcommand{\omwEnDefbertUnatt}{\attested{} + \glossbert{}}

\newcommand{\omwEnDefbertOwt}{\owt{} + \glossbert{}}
\newcommand{\omwEnDefbertAwt}{\awt{} + \glossbert{}}

\usepackage{comment}
\usepackage{graphicx}

\title{UAlberta at SemEval 2022 Task 2: 
Leveraging Glosses and Translations for Multilingual Idiomaticity Detection}

\author{
 Bradley Hauer,
 Seeratpal Jaura,
 Talgat Omarov,
 Grzegorz Kondrak\\
 Alberta Machine Intelligence Institute, Department of Computing Science\\
 University of Alberta, Edmonton, Canada\\
 {\tt {\{bmhauer,seeratpa,omarov,gkondrak\}}@ualberta.ca}
}

\begin{document}

\maketitle

\begin{abstract}
We describe the University of Alberta systems for
the SemEval-2022 Task 2 on multilingual idiomaticity detection.
Working under the assumption 
that idiomatic expressions are non-compositional,
our first method integrates information on the meanings 
of the individual words of an expression
into a binary classifier.
Further hypothesizing that literal and idiomatic expressions
translate differently,
our second method translates an expression in context,
and uses a lexical knowledge base to determine
if the translation is literal.
Our approaches are grounded in linguistic phenomena,
and leverage existing sources of lexical knowledge.
Our results offer support for both approaches,
particularly the former.
\end{abstract}

\section{Introduction}
\label{intro}

In this paper, we describe the University of Alberta systems for
the task of classifying multi-word expressions (MWEs) 
in context
as either \emph{idiomatic} or \emph{literal}
\cite{tayyarmadabushi2022}.
Each instance in the data includes 
a MWE (e.g., {\em closed book}), its language,
and its context, composed of the three surrounding sentences.
We participate in both the zero-shot and one-shot settings.
 
While the exact definitions of the two key terms are not
stated explicitly in the task description\footnote{\url{https://sites.google.com/view/semeval2022task2-idiomaticity}},
it is suggested that 
{\em idiomatic} is synonymous with {\em non-compositional}.
The Pocket Oxford Dictionary defines 
{\em idiomatic} as 
``not immediately comprehensible from the words used,''
and {\em literal} as 
``taking words in their basic sense.''
Therefore, we adopt the following MWE {\em compositionality criterion} 
\[
\mbox{literal} 
\equiv \mbox{compositional} 
\equiv \neg\,\mbox{idiomatic}
\]
where the three terms are considered to be Boolean variables.
In addition,
the shared task 
considers all proper noun MWEs 
(e.g., {\em Eager Beaver}) 
as literal.

Our goal is to explore the idea that
glosses and translations of word senses 
can help decide whether 
the meaning of a given MWE occurrence is compositional.
Based on the above-stated compositionality criterion, 
this in turn could facilitate idiomaticity detection.
In particular, we hypothesize that
at least one of the words in any idiomatic expression 
is used in a non-standard sense.
Following the intuition that
a traditional word sense disambiguation (WSD) system can only
identify senses that are included in a given sense inventory,
we propose two methods that 
indirectly detect non-standard senses by 
leveraging either glosses or translations of senses
from such an inventory.

\begin{figure}[t]
    \centering
    \includegraphics[width=0.48\textwidth]{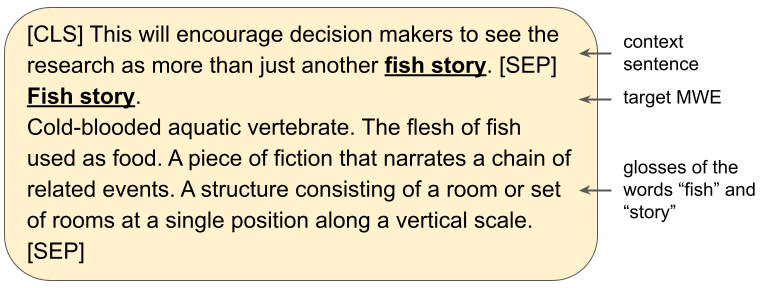}
    \caption{An example of \glossbert{} input.}
    \label{fish_story}
\end{figure}

Our gloss-based method follows from the intuition that
the meaning of a given MWE occurrence is related
to any of the existing sense glosses of its component words
{\em only if the expression is compositional}.
Therefore, 
the addition of the glosses to the context of the expression
should help the classifier in deciding whether the MWE is used
in a literal or idiomatic sense.
We implement this method by
adding the glosses of each sense of each individual word,
retrieved from a lexical knowledge base,
to the input to a neural classifier 
which fine-tunes multilingual BERT 
\cite[mBERT;][]{devlin2019} 
for the idiomaticity detection task.
We refer to this method as \glossbert{}
(Figure~\ref{fish_story}).

Our translation-based method follows from the observation
that compositional expressions
are typically translated word-for-word (``literally''),
which implies that
each content word and its translation should have the same meaning.
Therefore,
each such multilingual word pair 
should share a multi-synset in a multi-wordnet 
\cite{hauer2020set}.
The procedure is as follows:
(1) translate the MWE in context;
(2) word-align the source and target sentences;
(3) lemmatize and POS-tag the source MWE;
and (4) for each lemma in the MWE,
search for a multi-synset that contains both the lemma and its translation.
This method is unsupervised,
and we refer to it as \translation{}.

Our results provide evidence that 
leveraging lexical resources is beneficial for 
idiomaticity detection.
In particular,
our gloss-based method,
when combined with a type-based {\attested} heuristic,
is among the top-scoring submissions in the one-shot setting.
The heuristic
is based on the observation that some MWEs are inherently idiomatic or literal,
regardless of their context,
which is confirmed by our analysis of the development set annotations.

\section{Related Work}
\label{relwork}

Early attempts to represent idiomatic MWEs involve 
treating idiomatic phrases as individual tokens and 
learning corresponding static embeddings \citep{mikolov2013phrases}.
However, \citet{cordeiro2016} show that 
the effectiveness of this method is limited by 
data sparsity for longer idiomatic expressions. 
Furthermore, \citet{shwartz2019} and \citet{garcia2021} 
conclude that idiomaticity is not yet accurately represented
even by contextual embedding models.
\citet{madabushi2021astitchinlanguagemodels} 
create a new manually labeled dataset containing 
idiomatic and literal MWEs,
and propose a method based on a pre-trained neural language model. 

Regarding using lexical translations for idiomaticity detection, 
\newcite{moiron2006} measure semantic entropy in bitext alignment statistics,
while \citet{salehi-etal-2014-detecting} predict compositionality 
by presenting an unsupervised method that uses 
Wiktionary translation, synonyms, and definition information.
We extend these ideas 
by applying machine translation,
and consulting a multilingual lexical knowledge base.

Our prior work 
has already demonstrated the utility of lexical translations
for various semantic tasks,
including prior SemEval tasks on
predicting cross-lingual entailment \cite{hauer2020semeval}
and contextual synonymy detection \cite{hauer2021semeval},
as well as
word sense disambiguation \cite{luan2020},
and homonymy detection \cite{hauer2020ohpt, habibi2021}.

\section{Methods}
\label{methods}

In this section, we describe our methods for \task{}.

\subsection{Baseline mBERT}
\label{baseline}
We re-implemented the mBERT classifier baseline
\citep{devlin2019} 
following the methodology of \citet{madabushi2021astitchinlanguagemodels}.
The model takes the context sentence and the relevant MWE as an input,
and outputs a binary label indicating the idiomaticity of the target MWE.
The input sequence is constructed by concatenating the MWE 
to the end of the context sentence after the special [SEP] token.

It is important to note the differences between our re-implementation
and the official baseline provided by the task organizers.
In the official baseline, the organizers add the target MWE as an additional
feature in the one-shot setting but not in the zero-shot setting. 
Furthermore, the organizers include the sentences preceding and succeeding 
the target sentence only in the zero-shot setting. In our re-implementation,
we add the target MWE and exclude the preceding and succeeding sentences 
in both zero-shot and one-shot settings.

\subsection{Gloss-based Method}
\label{glossbert}

Our first method, \glossbert{},
extends the baseline model by adding the glosses 
of all possible senses of each individual word
in the target MWE to the classifier's input.
The intuition
is that 
the addition of the glosses to the input should help the classifier 
decide if the meaning of the target MWE can be deduced 
from the definitions of the individual words, 
i.e., if it is compositional.
In the example in Figure~\ref{fish_story}, 
the disparity between the context in which \emph{fish story} appears, 
and the glosses of the various senses of the words \emph{fish} and \emph{story} 
indicates that the MWE is idiomatic in this context.

The intuition for this method is that
non-native speakers can identify idiomatic expressions,
provided they understand the standard meanings of the words which comprise them.
Suppose that the vocabulary of a
non-native speaker covers most of the essential
words necessary to understand a language, but not idiomatic expressions. 
Even if the speaker cannot deduce the meaning of an idiomatic expression 
in context, they can guess that the expression
was used in an idiomatic sense because individual
words of this expression do not make sense in the given context.

\subsection{Translation-based Method}
\label{translation}

\begin{comment}
\begin{algorithm}[t] 
  \begin{algorithmic}[1]
  \scriptsize
  \Require{source sentence and MWE} 
  \Ensure{binary classification of MWE as idiomatic/literal}
  \Statex
  \Function{BabelNetQuery}{mwe} 
      \If{properNoun(mwe)}
          \State \Return{Literal}
      \EndIf
      \If {DoubleQuotes(mwe)}
          \State \Return{Idiomatic}
      \EndIf

      \State compositional := 0
      \For{each language l}
          \State words := 0
          \For{each word w in mwe}
              \State t = translation(L, w)
              \If{sharedSynsetsOMW(w, t) or sharedSynsetsBN(w, t)}
                  \State words = words + 1
            \EndIf
          \EndFor
          \If{$words > threshold1$}
              \State compositional:= compositional + 1
          \EndIf
      \EndFor
      \If{$compositional >= threshold2$}
          \State \Return{Literal}
      \EndIf
      
      \State \Return{Idiomatic}
  \EndFunction
  \end{algorithmic}
\caption{\translation{}}
\label{translation_pcode}
\end{algorithm}
\end{comment}

Our \translation{} method
is based on translating the target MWE in context,
and leverages multilingual semantic resources.
The intuition behind this method is that idioms 
are generally specific to a particular language,
and, being non-compositional, their meanings cannot be conveyed
simply by translating the individual words.

Under this hypothesis, to classify an MWE as literal or idiomatic,
we need only determine whether the words in the MWE are translated literally.
We do this by
first identifying the translation of each word via alignment.
We then consult a multilingual wordnet, or \emph{multi-wordnet},
a lexical knowledge-base which
organizes words in two or more languages into multilingual synonym sets,
or \emph{multi-synsets}.
Each multi-synset corresponds to a unique concept,
and contains the words which express that concept.
Given a word in context, and a translation of that word in that context,
we consider the word to be literally translated
if it shares at least one multi-synset with its translation.

For example, consider an instance 
in which the MWE \emph{wedding anniversary}
is translated into Italian as {\em anniversario di matrimonio}.
Our method checks if either of the translation pairs
(\emph{wedding}, \emph{matrimonio})
and (\emph{anniversary}, \emph{anniversario}) 
share a multi-synset in a multi-wordnet.
We test two versions of this method:
in {\awt}, this condition must be satisfied for all content words in the MWE;
in {\owt}, detecting a literal translation for one word is sufficient
to classify the MWE as literal.
In addition, multiple languages of translation may be considered.

\subsection{Additional Heuristics}
\label{heuristics}

The annotation methodology for this shared task includes
proper nouns in the literal class.
We therefore use a part-of-speech tagger to detect proper nouns;
if any word in the MWE is tagged as a proper noun,
\translation{} automatically classifies it as literal
without further consideration.

In the one-shot setting, we also use a type-based heuristic
which we refer to as \attested{}.
The intuition behind this heuristic is
that certain MWEs are inherently idiomatic or literal,
regardless of the context that they appear in.
If the training data has no example of an MWE in a particular class,
the heuristic exploits this fact as evidence that the MWE 
should always be classified as the opposite, attested class.
For example,
this heuristic always classifies
{\em life vest} as idiomatic
and {\em economic aid} as literal,
as these are the only classes in which these MWEs appear 
in the training data.
In practice,
since \attested{} returns no classification
if the training set contains instances that belong to either class,
this heuristic must be used in combination with another method.

\subsection{Combination}
\label{combination}

Our \glossbert{} and \translation{} methods take different views of the data,
with the former using a neural language model and gloss information,
and the latter using translation and a lexical knowledge base.
We therefore consider combining the two methods.
In this approach, we independently apply \glossbert{} and \translation{}
to a given instance.
If the two methods agree, we return the agreed-upon classification;
if they disagree, we return a default class, which is a tunable parameter.
As with the other methods,
we can combine this method with the \attested{} heuristic
in the one-shot setting.

\section{Experiments}
\label{experiments}

We now describe our experiments,
including the tools and resources,
the experimental setup, 
the results, and a discussion of our findings.

\subsection{Lexical Resources}  
\label{lexres}

As lexical resources for sense translations and glosses,
we use two different multi-wordnets:
BabelNet \cite[BN;][]{navigli2010, navigli2012},
and Open Multilingual WordNet \cite[OMW;][]{bond2013}.
The \glossbert{} method and the alignment tool access 
BN 4.0 via the provided Java API\footnote{\url{https://babelnet.org/guide}}. 
For the \translation{} method, we access the BN 5.0 via the HTTP API. 
We access OMW via the NLTK interface 
\cite{nltk}.
For the \translation{} method, 
we consider the translation of a word to be literal 
if it shares a multi-synset with the word in either BN or OMW.
For lemmatization and POS tagging,
we use TreeTagger\footnote{We use the pre-trained models 
for English, Portuguese, and Galician
from \url{https://cis.uni-muenchen.de/~schmid/tools/TreeTagger.}}
\cite{schmid2013}.

Both BN and OMW contain English glosses for most concepts,
but the availability of glosses in other languages varies.
In particular, OMW contains no Portuguese or Galician glosses.
With BabelNet, we experimented with two techniques:
using English glosses for all languages,
and
using glosses from the language of the instance, 
i.e. the source language, when available.
We refer to these variants as {\bnEnDefbert} and {\bnDefbert}, respectively.
Since \glossbert{} uses a multilingual pre-trained language model,
it can seamlessly handle input from multiple languages.
Furthermore, because of 
the relatively poor coverage of Galician in the lexical resources
(only 54\% of glosses are available in this language),
we attempt to leverage its close relationship to Portuguese
by processing Galician as if it was Portuguese.

\subsection{Translation and Word Alignment}

We translate the context sentence of each MWE 
with Google Translate API\footnote{\url{https://cloud.google.com/translate}}.
We translated English instances into Italian, and 
Portuguese/Galician instances into English,
because of the good coverage of these languages in our resources.
We also conducted development experiments 
with translation into less related languages,
as well as with combining translation information from multiple languages,
but we observed no consistent improvements.

We align each input sentence with its translation
using BabAlign \cite{luan2020},
which consults BabelNet 
to refine the alignments generated by a base aligner, 
FastAlign \cite{dyer2013}.
To further improve the alignment quality,
we augment the set of sentence-translation pairs
with additional parallel data 
from the OpenSubtitles parallel corpus
\cite{lison2016}.
We note that the English-Galician bitext 
is less than 1\% of the size of the other two bitexts.

\subsection{mBERT and \glossbert}

We fine-tune the mBERT-based models using the binary classification objective 
on the labeled training dataset.
In the zero-shot setting, the MWEs in the training data 
are disjoint from those in the development and test splits,
while in the one-shot setting, all MWEs in the development and test splits
have at least one example in the training data.
In the zero-shot setting, we trained the models 
only on the zero-shot training set,
while
in the one-shot setting, we trained the models on both 
training sets.
In particular, we fine-tuned the models for 20 epochs 
with a maximum sequence length of 256, a learning rate of 2e-5,
and a per device batch size of 16, using the HuggingFace Transformers
library.\footnote{\url{https://huggingface.co}}

\begin{table*}[t]
    \centering
    \small
     \begin{tabular}{|c|l|cc|cc|cccc|cccc|}
        \hline
        & & \multicolumn{4}{c|}{Development results} 
        & \multicolumn{8}{c|}{Test results} \\  
        \cline{3-14}
        & & \multicolumn{2}{c|}{Zero-Shot} & \multicolumn{2}{c|}{One-Shot} 
        & \multicolumn{4}{c|}{Zero-Shot} & \multicolumn{4}{c|}{One-Shot}\\ 
        \cline{3-14}
        &
	& EN & PT & EN & PT & 
        EN & PT & GL & ALL & EN & PT & GL & ALL \\
        \hline
        0 & Baseline & 66.2 & 63.9 & 87.0 & 86.7 & 
        70.7 & 68.0 & 50.7 & 65.4 & 88.6 & 86.4 & 81.6 & 86.5 \\
        \hline
        1 &  \baseline{} & 74.6 & 62.5 & 85.7 & 85.9 & 
        \textbf{75.1} & 63.3 & \textbf{61.1} & 68.2 & 90.0 & 83.6 & 86.6 & 87.7 \\
        \hline
        2 & \bnDefbert{} & \textrm{75.5} & 64.8 & 85.4 & 86.7 & 
        72.0 & 66.4 & 57.8 & 67.2 & \textbf{95.7} & 88.5 & 88.9 & 92.2 \\
        \hline
        3 & \bnEnDefbert{} & 75.3 & \textrm{66.4} & 87.6 & 86.6 & 
        73.4 & \textbf{68.4} & 59.7 & \textbf{69.5} & 95.0 & \textbf{89.3} & 87.9 & 91.8 \\
        \hline
        4 & \omwEnDefbert{} & 74.8 & 64.5 & 87.1 & 84.5 & 
        71.0 & 65.6 & 56.5 & 66.5 & 92.4 & 86.7 & 88.5 & 90.1 \\
        \hline
        5 & \omwEnDefbertUnatt{} & - & - & {\bf 92.0} & {\bf 87.7} & 
        - & - & - & - & 94.5 & 89.2 & \textbf{91.2} & \textbf{92.4}\\
        \hline
        6 & \omwEnDefbertOwt{}  & {\bf 77.3} & 64.9 & 84.5 & 78.0 & 
        68.2 & 54.6 & 56.3 & 62.7 & 85.9 & 70.6 & 78.2 & 80.6 \\
        \hline
        7 & \omwEnDefbertAwt{} & 66.4 & {\bf 69.2} & 73.7 & 78.0 & 
        65.4 & 62.5 & 54.3 & 62.1 & 80.3 & 73.8 & 73.9 & 77.3\\
        \hline
    \end{tabular}
    \caption{
        The macro F1 scores on the development and test datasets. 
        Our official submissions are in rows 4-7.
        Where not otherwise specified, 
        \glossbert{} is in the OMW-en configuration.
    }
    \label{tab:results}
\end{table*}

\subsection{Development experiments}

Table~\ref{tab:results} contains the results of the following models:
the official mBERT-based baseline (row 0)
as reported by the shared task organizers,
our re-implementation of the official baseline (row 1),
three variants of \glossbert{} method 
which is based on {\baseline} (rows 2-4),
\glossbert{} combined with the \attested{} heuristic (row 5),
and the \translation{} method combined with \glossbert{} 
(rows 6-7)\footnote{After the test output submission deadline, 
we discovered errors in our implementation of the \translation{} methods. 
We report our original results for consistency with the official results.}.
For rows 1-5 we average the macro F1 score obtained 
over five runs with random initializations.

Our experiments with \glossbert{} 
explored the impact of adding glosses to the {\baseline} model,
including the source and language of the glosses.
With English glosses retrieved from BabelNet, \glossbert{}  
improves the total score over the {\baseline} model
in the zero-shot setting,
especially on Portuguese.
The results also suggest that the English glosses 
may be preferable to glosses in the source language,
a finding which could simplify work on lower-resourced languages,
where glosses may not be available.

Combining the predictions of the mBERT-based 
models with the \attested{} heuristic
improves the one-shot F1 scores
in all cases
(row 5 vs. row 4). 

The \translation{} methods achieve the best results 
when combined with \glossbert{}
on the development set in the zero-shot setting:
{\owt} for English (row 6), 
and {\awt} for Portuguese (row 7).
This demonstrates the utility of 
using lexical translation information for idiomaticity detection
when 
annotated training data is not available.

\subsection{Error Analysis}

We found that the \glossbert{} method performs slightly better,
by about 1\% F1,
on literal instances as compared to idiomatic instances
in the one-shot setting.
In other words, the method is less likely to make an error
when given a literal instance.
We speculate that this is explained by 
the model's consistent classification of
proper nouns as literal expressions.
Indeed, 
a proper noun is identified incorrectly in only one instance.
The fraction of idiomatic vs.\ literal instances 
is 39\% in English and 56\% in Portuguese.

For the \translation{} method, 
a large number of of errors were caused by 
a literal translation of an idiomatic expression 
by Google Translate,
even though the corresponding expression
is not meaningful in the target language.
For example,
``she was different, like a \underline{closed book}''
is translated into Italian as
``era diversa, come un \underline{libro chiuso}''
even though the Italian translation 
does not carry the meaning of a person being secretive. 
In a few cases, the translation would simply 
copy the source language expression,
yielding output which is not fully translated.
In addition, some correct lexical translations are not in our
lexical resources.
Finally, a number of incorrect idiomatic predictions could be traced
to word alignment errors,
especially in cases of 
many-to-one alignments
(e.g., {\em bow tie} correctly translated as {\em papillon}).

Manual analysis performed on the development set 
corroborates our hypothesis that most multi-word expressions 
are inherently idiomatic (e.g., {\em home run})
or literal (e.g., {\em insurance company}). 
Only about one-third of the expressions are ambiguous 
in the sense that they can be classified as either class 
depending on the context (e.g. {\em closed book}).
Our judgements are generally corroborated by the gold labels,
with the exception of proper nouns, which are consistently marked as literal.
The {\attested} heuristic (Section~\ref{heuristics}),
which is based on this observation, 
obtains a remarkable 98.3\% precision and 55.8\% recall 
on the set of 739 instances in the development set.

\subsection{Test set results}

The results on the test set are shown in Table~\ref{tab:results}.
Our best results are produced by 
{\bnEnDefbert}
in the zero-shot setting,
and the combination of \glossbert{} with the \attested{} heuristic 
in the one-shot setting.
The latter also obtains the best result on Galician,
which demonstrates its applicability to low-resource languages,
as this method only requires English glosses.

The results of combining \glossbert{} with \translation{}
are well below the baseline,
which may be due to a different balance of classes in the test set,
omissions in lexical resources, 
and/or errors in our initial implementation.
Another possible reason is that 
modern idiomatic expressions
are often translated word-for-word (``calqued''),
especially from English into other European languages.
Examples from the development set include 
{\em flower child, banana republic}, and {\em sex bomb}. 

\section{Conclusion}
\label{conclusion}

Our top result
ranks third overall in the one-shot setting.
The corresponding method is applicable to a wide variety of languages.
It takes advantage of the ability of neural language models
to seamlessly incorporate textual information such as glosses,
even if it is expressed in a different language.
These results strongly support our hypothesis that
the gloss information of individual words
can improve idiomaticity detection.
Moreover, our development results 
support the hypothesis
that non-compositional expressions can be identified through their translations.
These findings conform with our prior work on
leveraging translation for various semantic tasks 
(Section~\ref{relwork}).
We hope that this work will 
motivate further investigation
into the role of multilinguality in semantics.

\section*{Acknowledgments}

This research was supported by
the Natural Sciences and Engineering Research Council of Canada (NSERC),
and the Alberta Machine Intelligence Institute (Amii).

\bibliography{wsd}

\end{document}